\documentclass[submission,copyright,creativecommons]{eptcs}
 
\usepackage{algorithm}
\usepackage{pgfplots}
\usepackage{algpseudocode}
\usepackage{graphicx}
\usepackage{mathpartir}
\usepackage{wasysym}
\usepackage{enumitem}
\usepackage{txfonts}

\usepackage{amsthm}

\usepackage{amsmath}
\newtheorem{mydef}{Definition}
\newtheorem{prop}{Proposition}

\newtheorem{myex}{Exemple}

\title{An ASP-based Approach for Attractor Enumeration in Synchronous and Asynchronous Boolean Networks \\}

\author{Tarek Khaled and Bela\"id Benhamou\\
         \institute{Aix Marseille University,\, University of Toulon, CNRS, LIS, Marseille, France.}
         \email{\{tarek.khaled,belaid.benhamou\}@univ-amu.fr }}
         
\def\titlerunning{An ASP-based Approach for Attractor Enumeration in Boolean Networks}
\def\authorrunning{Tarek Khaled and Bela\"id Benhamou}

\begin{document}
\maketitle

\begin{abstract}
Boolean networks are conventionally used to represent and simulate gene regulatory networks. In the analysis of the dynamic of a Boolean network, the attractors are the objects of a special attention.  In this work, we propose a novel approach based on Answer Set Programming (ASP) to express Boolean networks and simulate the dynamics of such networks. Our work focuses on the identification of the attractors, it relies on the exhaustive enumeration of all the attractors of synchronous and asynchronous Boolean networks. We applied and evaluated the proposed approach on real biological networks, and the obtained results indicate that this novel approach is promising.
\end{abstract}

\section{Introduction}\label{sec1}

A gene regulatory network is a collection of genes interacting with each other. Each gene contains information determining its function. It is a specific biological system that represents how the genes interact in a cell for its survival, reproduction, or death. Among the approaches that are used to model these networks \cite{dejong2002}, we can find qualitative ones. These approaches allow for the capture of the most important properties, like the attractors which represent the sets of states to which the system converges.

Our goal in this work is to develop an exhaustive approach to analyze the dynamics of Boolean networks . We are dealing with two kinds of problems: finding all the possible stable states and enumerating all the stable cycles of a dynamic system. We apply two update modes that are the synchronous and asynchronous modes and use the ASP framework  to represent and solve the aforementioned problems. We will see that the identification of the attractors in a transition graph representing the dynamics of a Boolean network amounts to calculating and enumerating the stable models of the logical program expressing the interactions between the entities of the network. It is than important to have an efficient ASP solver that is able to enumerate the models in a reasonable time. In this work, we use the method introduced in \cite{method2018}.  This method relies on a Boolean enumeration process defined for the ASP paradigm according to the semantic introduced in \cite{newsem}. 

In what follows, we start by summarizing some notions on Boolean networks, then show in Section \ref{sec3},  how to express  gene networks as logic programs. In Section \ref{sec5}, we evaluate our approach on biological networks. We conclude the work in Section \ref{sec6}.


\section{Boolean Networks (BN)}

Le $V = \{v_1,..., v_n \}$ be a finite set of Boolean entities $v_i \in \{0,1\}$. A  system configuration $x = (x_1,\dots , x_n)$  is the assignment of a truth value $x_i \in \{0, 1\}$ to each element of $V$. The set of all configurations \cite{jacob1961}, also called \emph{the space of configurations}, is denoted by $X=\{0,1\}^n$. 

The dynamics of such a Boolean system is modeled by a  \emph{global transition function} $f$ and by an updating mode that define how the elements of $V$ are updated over time. Formally, we have : $f : X \rightarrow X$  such that $x = (x_1 ,\dots, x_n)$ $\mapsto$ $f(x) = (f_1(x), \dots, f_n(x))$, where $f_i : X \rightarrow \{0,1\}$ is \emph{a local transition function} that gives the evolution of the gene $x_i$ along time. The dynamic of a Boolean network is naturally described by a transition graph $TG$ that  is characterized by a transition function $f$ and an update mode. Formally:

\begin{mydef}
Let $X=\{0,1\}^n$ be the configuration space of a Boolean network and  $f : X \rightarrow X$ its associated global transition function. The transition graph  representing the dynamic of $f$ is the oriented graph $TG(f)=(X,T(f))$ where the set of vertices   is the configuration space $X$ and the set of arcs is 
$T(f) = \{(x,y) \in X^2 \;|\; x \neq y, x=(x_1 ,\dots,x_i,\dots, x_n), y=(f_1(x) ,\dots, f_i(x),\dots, f_n(x))\}$
\end{mydef}




There exist several update modes, among them the synchronous and the asynchronous modes. The synchronous update means that all the components of a configuration  $x = (x_1,\dots , x_n)$  are updated at the same time. Conversely, the asynchronous mode is an update in which only one component of $x$ is updated at each time.

An orbit in $TG(f)$ is a sequence of configurations $(x^0,x^1,x^2,...)$ such that either $(x^t,x^{t+1}) \in T(f)$ or $x^{t+1}=x^t$ when there is no successors for $x^t$. A cycle of length $r$ is a sequence of configurations $(x^1,\dots,x^r,x^1)$ with $r \geq 2$ whose configurations $x^1,\dots, x^r$ are all different. We can now give the meaning of an attractor in a dynamical system.  A configuration  $x = (x_1,\dots , x_n)$ of the transition graph $TG(f)$ is a stable configuration when  $\forall x_i\in V, x_i = f_i(x)$, thus $x=f(x)$. A stable configuration $x = (x_1,\dots , x_n)$ forms a trivial attractor of $TG(f)$. A sequence of configurations $(x^1,x^2,\dots,x^r,x^1)$ forms a stable cycle of $TG(f)$ when   $ \forall t<r, x^{t+1}$ is the unique successor of $x^t$ and $x^1$ is the unique successor of $x^r$. A stable cycle in $TG(f)$ forms a cyclic attractor.



Transition graphs represent an excellent tool for studying the dynamic behavior of an update function corresponding to a Boolean  network. However, in practice, biological data comes from experiments that generally give only correlations between the genes, but nothing on the dynamic of the network. The interaction graph is a static  representation of the regulations between the genes.  Each node $v_i$ of the interaction graph is a Boolean variable that represent the state of gene $i$ in the network. More precisely, if $v_i=1$ (resp. $v_i=0)$), then the gene $i$ is active (resp. inactive).  A positive (resp. a negative) arc  $(v_i,+,v_j)$ (resp. $(v_i,-,v_j)$) defined from the node $v_i$ to the node $v_j$ means that the gene $i$ is an activator (resp. or an inhibitor) of the gene $j$. In the following, when there is no confusion we will lighten the notation by simply writing $ i $ to express the  vertex $ v_i $.

\begin{mydef}
An interaction graph is a signed-oriented graph $IG = (V, I)$ where $ V = \{1,\dots, n \} $ is its set of vertices and $ I \subseteq V \times \{+, - \} \times V $ its set of signed arcs.
\end{mydef}



\begin{myex}\label{ex_G}
Consider a set of two genes $V=\{1,2\}$, the space configuration $X=\{0,1\}^2$ and the transition function $f$ defined as $f(x_1,x_2)=(x_2,x_1 \land \lnot x_2)$. From $f$, we derive the transition graphs corresponding to both synchronous and asynchronous update modes. The synchronous and asynchronous transition graphs are given in Figure \ref{fig_G}-{b} and Figure \ref{fig_G}-{c}. Figure \ref{fig_G}-{a} shows the interaction graph $IG = (V,I)$  where $V=\{1,2\}$ and $I=\{(1,+,2),(2,+,1),(2,-,2)\}$, corresponding to both previous transition graphs.



\begin{figure}[htb!]
\centering
  \begin{minipage}{.3\textwidth}
	\includegraphics[scale=.4]{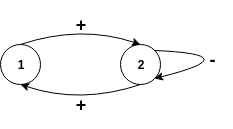}
	\caption{(a)}
  \end{minipage}
  \begin{minipage}{.2\textwidth}
	\includegraphics[scale=.4]{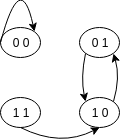}
	\caption{(b)}
  \end{minipage}
  \begin{minipage}{.2\textwidth}
	\includegraphics[scale=.4]{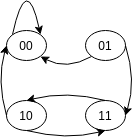}
  	\caption{(c)}
  \end{minipage}
  \caption{The synchronous (b) / asynchronous (c) transition graphs of a Boolean network represented by an interaction graph (a) having  two genes}
  \label{fig_G}
\end{figure}


The synchronous graph has two attractors that are the stable configuration $(0,0)$ and the stable cycle formed by the configurations $\{(1,0),(0,1)\}$. On the other hand, the asynchronous graph has only one attractor represented by the stable configuration $(0,0)$. We can also remark that the asynchronous graph contains a non-stable cycle $\{(0,1),(1,0)\}$. This last cycle is not stable  because there is an arc going out of it from $(0,1)$.
\end{myex}

\section{Using the ASP framework for Boolean network modeling and attractor computation}\label{sec3}

\subsection{Interaction graph modeling}

In this section, we show how to express the interaction graph associated with a Boolean network as an extended logic program $P_{IG}$. In other words, we represent the global transition function associated with the corresponding transition graph. The dynamics of the network will be represented by the answer sets of the logic program considered. We start with the rule ($r_1$) that encodes the notion of discrete time: 

$r_1: time(0..t). $ 

To compute the different configuration sequences of a given Boolean network, we have to observe its behavior under certain initial state conditions. This could require defining various combinations for the initial state. The number of possible combinations for the initial state could be very high. This renders the task very heavy for a hand user. We then decided to automate the process. To do this, we introduced the rules ($r_2$ and $r_3$) that will generate all the combinations of the initial state:
\\
$
r_2: v_i(0)  \leftarrow not \, \lnot v_i(0). \qquad
r_3: \lnot v_i(0) \leftarrow not \, v_i(0).
$

These rules force the solver to make choices for each gene, either it is active or inactive. Indeed, the absence of $v_i(0)$ makes $\lnot v_i(0)$ true, and conversely absence of $\lnot v_i(0)$ makes $v_i(0)$ true. In this way, different answer sets are automatically generated for each possible starting combination.

The rules from $\{r_4,r_5,r_6,r_7\}$ encodes the influences of one gene on another. That is, the activation or inhibition of a gene by an other gene.
\\
$
r_4: v_j(t+1) \leftarrow v_i(t) \qquad
r_5: \lnot v_j(t+1) \leftarrow \lnot v_i(t) \\
r_6: v_j(t+1) \leftarrow \lnot v_i(t) \qquad
r_7: \lnot v_j(t+1) \leftarrow v_i(t)
$

Rules $r_4$ and $r_5$  mean the following:  if the gene $v_i$ is active (resp. inactive) at time step $t$, then it will activate (resp. will inhibit) the gene $v_j$ at time step $t+1$ . These two rules represent the positive oriented arc $(v_i,+,v_j)$ of the associated interaction graph. Both rules $r_6$ and $r_7$ express the fact  that the activation (resp. inhibition) of the gene $v_i$ at time step $t$ will inhibit (resp. will activate) the gene $v_j$ at time step $t+1$. These rules encode the negative oriented arc $(v_i,-,v_j)$ of the interaction graph.  

The rules $r_8$ and $r_9$ are inertia rules that express what happens to a gene when there is no change at the next time step. That is, a gene preserve its state unless it was changed:
\\
$
r_8: v_i(t+1) \leftarrow v_i(t), \, not \; \lnot v_i(t+1)  \qquad
r_9: \lnot v_i(t+1) \leftarrow \lnot v_i(t), not \; v_i(t+1) 
$

In what follows, we will present some rules to manage Boolean networks where a given gene could have several interactions with the other genes. The main idea, is to express each local transition functions $f_i$  as a set of rules. To this end, we assume that any function $f_i$ is given in disjunctive normal form (DNF). Given the configuration $v=(v_1,v_2,...,v_j,...,v_n)$,  for each  node $v_i \in V$ of the interaction graph, we express its corresponding function $f_j$ by the following DNF formula:

$v_j(t+1)=f_j(v(t))= \bigvee\limits_{i=1}^{l_j}m_i^j$ , 
where  $m_i^j =(\pm v_{i_1}\land \pm v_{i_2}\land,\dots,\land \pm v_{i_k})$ 
and $i_h\in\{1..n\}$ $\forall h\in \{1..k\}$
The formula $m_i^j$ is a conjunction of literals representing positive / negative interactions of genes $v_{i_h}$ acting on $v_j(t)$. 
Let $DNF(\neg f_j(v(t))=\neg (\bigvee\limits_{i=1}^{l_j}m_i^j)=\bigvee\limits_{i=1}^{r_j}{{m}_i^{\prime j}}$ be the DNF form of $\neg f_j(v(t))$. 
The set of rules that encodes each function  $f_j$ is defined as follows:
\\
$
r_{10}: \{v_j(t+1) \leftarrow m_i^j(t) \; | \; 1 \leq i \leq l_j \},  j\in \{1,\dots,n\}\\
r_{11}: \{\neg v_j(t+1) \leftarrow {m'}_i^j(t) \; | \; 1 \leq i \leq r_j \},  j\in \{1,\dots,n\} 
$

The formula ${m'}_i^j$ is a conjunction of literals representing positive / negative interactions of genes  acting on $\neg v_j(t)$.

\begin{myex}
Consider the interaction graph given in Example \ref{ex_G}. The sets of rules generated by the rules $r_{10}$ and $r_{11}$ when applied to the considered interaction graph are the following:

$r_{10}: \{1(t+1) \leftarrow 2(t); \qquad 2(t+1) \leftarrow 1(t), \; \lnot2(t)\}$

$r_{11}: \{\neg 1(t+1) \leftarrow \neg 2(t); \qquad \neg 2(t+1) \leftarrow \neg 1(t); \qquad \neg 2(t+1) \leftarrow 2(t)\}$ 
\end{myex}

The rules $r_{10}$ and $r_{11}$ are applicable only for synchronous update mode. For the asynchronous update mode, we have to consider only one local transition function $f_j$ at each time step. To do this, we introduce a new predicate $Block(v_i,t)$ stating that $v_i$ is blocked for update at time step $t$. Obviously, each unblocked local transition can be performed.

We also add the rules $r_{12}$ and $r_{13}$ which express the fact that the state of the gene $v_i$ is updated each time its state at step $t+1$ is different from its state at the previous step $t$.
$
r_{12}: Change(v_i,t) \leftarrow v_i(t+1), \lnot v_i(t)  \qquad
r_{13}: Change(v_i,t) \leftarrow \lnot v_i(t+1), v_i(t) 
$


To enforce the asynchronous mode, we establish the rule $r_{14}$ to allow only one possible local transition and block all the others. This rule says that if a given $v_i$ is not blocked and its also updated then all the other $v_j$ will be blocked. For the asynchronous update mode we reconsider both rules $r_{10}$ and $r_{11}$ by involving the new predicate $Block(v_i,t)$ and obtain the rules $r_{15}$ and $r_{16}$. This rules state that a local transition can be made unless it is blocked. In other words, the local transition function $f_i$ is used to update $v_j(t)$ if $Block(v_j,t)$ is not true.
\\
$
r_{14}: \{\{Block(v_k,t)\} \leftarrow Change(v_i,t), not \; Block(v_i,t) \; | \forall \; k \in \{1,..., n\} \backslash \{i\} \}. \\
r_{15}: \{v_j(t+1) \leftarrow m_i^j(t), not \; Block(v_j,t) \; | \; 1 \leq i \leq l_j \},  j\in \{1,\dots,n\}\\
r_{16}: \{\neg v_j(t+1) \leftarrow {m'}_i^j(t), not \; Block(v_j,t) \; | \; 1 \leq i \leq r_j \},  j\in \{1,\dots,n\} 
$

\begin{myex}
The sets of rules generated by the rules $r_{14}$, $r_{15}$, and   $r_{16}$ when applied to the interaction graph of Example \ref{ex_G} are the following:\\  
$r_{14}: \{Block(1,t) \leftarrow Change(2,t) ,not \; Block(2,t); \hspace{0.2cm} Block(2,t) \leftarrow Change(1,t) , not \; Block(1,t)\}$ \\
$r_{15}: \{1(t+1) \leftarrow 2(t), not \; Block(1,t); \hspace{0.2cm} 2(t+1) \leftarrow 1(t), \; \lnot2(t), not \; Block(2,t)\}$ \\
$r_{16}: \{\neg 1(t+1) \leftarrow \neg 2(t), not \; Block(1,t); \hspace{0.2cm} \neg 2(t+1) \leftarrow \neg 1(t), not \; Block(2,t); \hspace{0.2cm} \neg 2(t+1) \leftarrow 2(t),not \; Block(2,t)\}$ 
\end{myex}

The rules explained above form the logic program $P_{IG}$ representing the interaction graph of the considered gene network. Now, we are able to establish the correspondence between an answer set of $P_{IG}$ and a a sequence of configurations  of the corresponding transition graph.

\begin{prop} \label{prop1}
Let $P_{IG}$ be the logic program representing the interaction graph $IG$ having a global transition function $f$ and $TG(f)$ the corresponding transition graph. A tuple $x=(x^0,\dots,x^t)$ is a sequence of configurations of $TG(f)$ , if only if  $I = \{(v_1(0),\dots,v_n(0)),\dots,(v_1(t),\dots,v_n(t))\}$ is an answer set of $P_{IG}$ such that the set of all the literals $(v_1(i),\dots,v_n(i))$ fixed at the step $i\in\{0,...,t\}$  corresponds to the state of the genes of the configuration $x^i\subseteq x$ defined at the step $i$ in the transition graph $TG(f)$.
\end{prop}


\subsection{The calculation of the attractors}

One of the methods used to analyze the dynamics of a Boolean network is to enumerate all the possible configurations and  run a simulation from each of them. The method enumerates all the possible state sequences of the transition graph. This ensures that all the attractors will be detected. In this approach, we search for all the sequences of configurations  a given length $n$ in the transition graph of a Boolean network. We say that a sequence has length $n$ if it has $n$ transitions. When a sequence of states is found, we check if it contains a cycle. Since each state in a synchronous transition graph has a unique successor then when a sequence of states enter in a cycle, it never leaves it. This means that each cycle in a synchronous transition graph is stable. However, in the case of asynchronous transition graphs, the states could have multiple transitions. Thus, in general,  the cycles are not necessarily stable. There may exist stable cycles and unstable cycles.


We can determine the presence of a cycle in a sequence of states by checking whether the last state  occurs at least twice in the path corresponding to the considered sequence. Clearly, all the states between any two occurrences of the last state belong to a cycle.  For stable configuration detection, it is sufficient to check whether the successor state is the same as the last one. Since we can have an exponential number of possible states in a transition graph, an explicit enumeration of all the states is cumbersome for large networks. We want to avoid this exhaustive enumeration performed by the naive simulation of the network dynamics. To do this, we keep track of the cycles already found to eliminate them in the next iterations. If for a given sequence length, we do not find any cycle, the algorithm doubles the value of $n$ and  searches for a sequence having the new length $2n$. The algorithms stop when no sequence of configurations is found. It means that all the cycles are already found. Once all the cycles have been found, we can only find sequences shorter than the current length. The general schema of the proposed method is presented in Algorithm \ref{algo}.

\begin{algorithm}[htb!]
	\footnotesize
    \caption{The general schema of the cycle search algorithm }
  \begin{algorithmic}[1]
	\Require $P_{IG}$: the logic program representing the interaction graph
	\State I=ASP-Solver($P_{IG}$)
	\While {I is a new answer set of $P_{IG}$}
	\State attractor\_is\_found = False
	\State $x_I=(x^0,x^1,..., x^t)$ is the sequence of configurations corresponding to I 
	\State $i=t$
	\While {(($i \geq 0$) and not (attractor\_is\_found)) }
	\If {$x^t=x^i$} 
	\State attractor\_is\_found = True
	\State $attractors = attractors \cup \{x^{i+1},..., x^n \}$
	\State $P_{IG}$ = $P_{IG} \cup_{j\in\{i+1,t\}} \{ \leftarrow v_1(j), v_2(j),\dots,v_n(j)\}$
	\EndIf
	\State $i=i-1$
	\EndWhile 
    \If{ not(attractor\_is\_found)} 
	\State $n = 2*n$
	\EndIf
	\State I=ASP-Solver($P_{IG}$)
	\EndWhile
  \end{algorithmic}
  \label{algo}
\end{algorithm}


In what follows, we start by generating an extended logic program $P_{IG}$ representing the interaction graph according to the schema described in the previous sub-section.  The logic program is generated for $n$ time steps. We use the ASP system presented in \cite{method2018} based on the semantics introduced in \cite{newsem} to compute the answer sets representing the sequences of configuration of a particular length $n$ in the transition graph. If an answer set is found, the algorithm checks whether there is cycle or a stable configuration in the sequence corresponding to this answer set. In the affirmative case, we build and add some constraint rules for each of the cycles already to avoid them in the next steps found states. That is, By adding these added rules to the logic program $P_{IG}$,  eliminate all the answer sets that could contain an attractor already found. If the solver does not find any answer set, then no configuration sequence of length $n$ exists. This implies that all the cycles have been already identified. 

In the case of the synchronous update mode, each cycle correspond to a stable cycle. But for the asynchronous mode, the cycles of the transition graph  are not necessary stable. The could be unstable cycles. To detect the instability of a cycle, one can verify at each step of the cycle, if the current configuration could evolve  to a new configuration that is not a part of the cycle. In the affirmative case, we proved the instability of the cycle, otherwise the cycle is stable.

Now we will show how we can check if a given cycle is stable or unstable. 

\begin{prop}\label{prop2}
Let $P_{IG}$ be the logic program representing the interaction graph $IG$ having a global transition function $f$, $TG(f)$ the corresponding transition graph and $I = \{(v_1(0),\dots,v_n(0)),\dots,(v_1(t),\dots,v_n(t))\}$ is an answer set of $P_{IG}$ corresponding to the sequence of configuration $x_I$ in $TG(f)$. If a subset of literals $I_s=  \{(v_1(1),\dots,v_n(1)),\dots,(v_1(r),\dots,v_n(r)),(v_1(r+1),\dots,v_n(r+1))\}\subseteq I$ corresponding to a sequence of configurations $(x^1,\dots,x^r,x^1)\subseteq x_I$ forms a stable cycle in $TG(f)$, then every  answer set $J$ of $P_{IG}$ different from $I$ ($J\neq J$), is such that $J \cap I_s = \{ \varnothing \}$
\end{prop}

We can do the stability check by a slight modification in the ASP solver \cite{method2018} that we used to compute the answer sets. Indeed, for each answer set of the program $P_{IG}$ containing a cycle,  we have to check for each of its sub-set of literals $\{(v_1(i),\dots,v_n(i))\}$ corresponding to a configuration $x^i$ of the cycle, if a new sub-set of literals $\{(v_1(i + 1),\dots,v_n(i + 1))\}$ corresponding to a configuration $x^{i+1}$ different from the successor  of $x^i$ in the cycle can be deduced. To do this, we try to produce a different configuration at each choice point $not \; Block(v_i,t)$ of the branch corresponding to that answer set. This could be done by choosing a different choice point literal $not \; Block(v_j,t)$. We integrated this operation  in the resolution process of the method \cite{method2018} that we used to compute the answer sets of $P_{IG}$.

\section{Experimental Results}\label{sec5}

To demonstrate the validity of our approach on Boolean network simulation and attractor discovery, we applied it on real biological networks. We checked the method for both synchronous and asynchronous update modes on real genetic networks found in the literature.  We tested the method on the networks  {\it yeast cell cycle} \cite{ay2009} and {\it fission yeast cell cycle} studied in \cite{davidich2008}. We also applied the method on the network {\it T-helper cell differentiation} described in \cite{garg2007}. We computed all the attractors of these Boolean networks when using the synchronous and asynchronous modes. 
We  focus here  on the computation times and the number of attractors found. The obtained results are presented in Table \ref{tab_res}. We can see that the method performs relatively fast on all of the networks. We also observe that the attractors in the synchronous case often coincide with those in the asynchronous case. This is due to the presence  of a great number of stable configurations in comparison to the number of stable cycles in these networks. It is well known that the stable configurations are often the same in both synchronous and asynchronous update modes \cite{garg2008}.

\begin{table}[htb!]
\footnotesize
\centering
\begin{tabular}{lllll}
\hline
Network & Genes &	Attractors &	Update Schema &	Time(Sec) \\
\hline
Yeast cell cycle &	11 &	6 &	Synchronous &	2,21 \\
	&11&	6&	Asynchronous&	0,56 \\
Fission Yeast	&10	&11&	Synchronous	&1,82 \\
	&10&	12&	Asynchronous&	0,5 \\
Th cell differentiation &	23&	2&	Synchronous&	0,37 \\
	&23&	2&	Asynchronous&	0,43 \\
\hline
\end{tabular}
\caption{The results obtained on common graph regulatory networks found in the literature}
\label{tab_res}
\end{table}

The obtained results can not be compared to the ones of the method presented in \cite{mushthofa2014}. Indeed, with this method, the user must choose a specific semantic of activation on which the dynamic evolution will be based. There is two semantics. The first one  activates a gene when at least one of its activators is active and no inhibitor is active. In the second semantics, a gene is activated when it has more activators expressed than inhibitors. The chosen activation semantic is applied for all the genes of the model, while the activation rules in our method are specific to each gene and based on the transition functions.

\section{Conclusion}\label{sec6}

Boolean networks are a widespread modeling technique for analyzing the dynamic behavior of gene regulatory networks. By using Boolean networks, we can capture the network attractors, which are often useful for studying the biological function of a cell. We proposed a method dedicated to find stable configuration and cycles (stable and unstable) for a chosen update mode (synchronous or asynchronous). The advantage of our approach is the exhaustive enumeration thanks to the use of ASP framework and the ASP solver introduced in \cite{method2018}. The proposed approach is applied on real life regulatory networks, and the obtained results look promising  and the room for improvement is important. 

We plan to extend this work by considering adaptations and optimizations of the approach to address larger boolean networks. First, the elimination feature used to remove attractors already found could be improved. The technique that we use currently consist on adding rules to the logic program to exclude the attractors already found. This technique could be memory consuming when we deal with large networks. Another trail consists in generating and increasing the size of path in a more adaptive and insightful way. This could avoid us to traverse unnecessarily long paths.

\bibliography{aaai_bib}

\begin{thebibliography}{1}
\providecommand{\bibitemdeclare}[2]{}
\providecommand{\surnamestart}{}
\providecommand{\surnameend}{}
\providecommand{\urlprefix}{Available at }
\providecommand{\url}[1]{\texttt{#1}}
\providecommand{\href}[2]{\texttt{#2}}
\providecommand{\urlalt}[2]{\href{#1}{#2}}
\providecommand{\doi}[1]{doi:\urlalt{http://dx.doi.org/#1}{#1}}
\providecommand{\bibinfo}[2]{#2}

\bibitemdeclare{article}{ay2009}
\bibitem{ay2009}
\bibinfo{author}{Ferhat \surnamestart Ay\surnameend}, \bibinfo{author}{Fei
  \surnamestart Xu\surnameend} \& \bibinfo{author}{Tamer \surnamestart
  Kahveci\surnameend} (\bibinfo{year}{2009}): \emph{\bibinfo{title}{Scalable
  steady state analysis of Boolean biological regulatory networks}}.
\newblock {\sl \bibinfo{journal}{PloS one}}
  \bibinfo{volume}{4}(\bibinfo{number}{12}), p. \bibinfo{pages}{e7992},
  \doi{10.1371/journal.pone.0007992}.

\bibitemdeclare{article}{newsem}
\bibitem{newsem}
\bibinfo{author}{Belaïd \surnamestart Benhamou\surnameend} \&
  \bibinfo{author}{Pierre \surnamestart Siegel\surnameend}
  (\bibinfo{year}{2012}): \emph{\bibinfo{title}{A New Semantics for Logic
  Programs Capturing and Extending the Stable Model Semantics}}.
\newblock {\sl \bibinfo{journal}{Tools with Artificial Intelligence (ICTAI)}},
  pp. \bibinfo{pages}{25--32}, \doi{10.1109/ICTAI.2012.167}.

\bibitemdeclare{article}{davidich2008}
\bibitem{davidich2008}
\bibinfo{author}{Maria~I \surnamestart Davidich\surnameend} \&
  \bibinfo{author}{Stefan \surnamestart Bornholdt\surnameend}
  (\bibinfo{year}{2008}): \emph{\bibinfo{title}{Boolean network model predicts
  cell cycle sequence of fission yeast}}.
\newblock {\sl \bibinfo{journal}{PloS one}}
  \bibinfo{volume}{3}(\bibinfo{number}{2}), p. \bibinfo{pages}{e1672},
  \doi{10.1371/journal.pone.0001672}.

\bibitemdeclare{article}{dejong2002}
\bibitem{dejong2002}
\bibinfo{author}{Hidde \surnamestart De~Jong\surnameend}
  (\bibinfo{year}{2002}): \emph{\bibinfo{title}{Modeling and simulation of
  genetic regulatory systems: a literature review}}.
\newblock {\sl \bibinfo{journal}{Journal of computational biology}}
  \bibinfo{volume}{9}(\bibinfo{number}{1}), pp. \bibinfo{pages}{67--103},
  \doi{10.1089/10665270252833208}.

\bibitemdeclare{article}{garg2008}
\bibitem{garg2008}
\bibinfo{author}{Abhishek \surnamestart Garg\surnameend},
  \bibinfo{author}{Alessandro \surnamestart Di~Cara\surnameend},
  \bibinfo{author}{Ioannis \surnamestart Xenarios\surnameend},
  \bibinfo{author}{Luis \surnamestart Mendoza\surnameend} \&
  \bibinfo{author}{Giovanni \surnamestart De~Micheli\surnameend}
  (\bibinfo{year}{2008}): \emph{\bibinfo{title}{Synchronous versus asynchronous
  modeling of gene regulatory networks}}.
\newblock {\sl \bibinfo{journal}{Bioinformatics}}
  \bibinfo{volume}{24}(\bibinfo{number}{17}), pp. \bibinfo{pages}{1917--1925},
  \doi{10.1093/bioinformatics/btn336}.

\bibitemdeclare{article}{garg2007}
\bibitem{garg2007}
\bibinfo{author}{Abhishek \surnamestart Garg\surnameend},
  \bibinfo{author}{Ioannis \surnamestart Xenarios\surnameend},
  \bibinfo{author}{Luis \surnamestart Mendoza\surnameend} \&
  \bibinfo{author}{Giovanni \surnamestart DeMicheli\surnameend}
  (\bibinfo{year}{2007}): \emph{\bibinfo{title}{An efficient method for dynamic
  analysis of gene regulatory networks and in silico gene perturbation
  experiments}}, pp. \bibinfo{pages}{62--76}.
\newblock \doi{10.1007/978-3-540-71681-5\_5}.

\bibitemdeclare{article}{jacob1961}
\bibitem{jacob1961}
\bibinfo{author}{Fran{\c{c}}ois \surnamestart Jacob\surnameend} \&
  \bibinfo{author}{Jacques \surnamestart Monod\surnameend}
  (\bibinfo{year}{1961}): \emph{\bibinfo{title}{Genetic regulatory mechanisms
  in the synthesis of proteins}}.
\newblock {\sl \bibinfo{journal}{Journal of molecular biology}}
  \bibinfo{volume}{3}(\bibinfo{number}{3}), pp. \bibinfo{pages}{318--356},
  \doi{10.1016/S0022-2836(61)80072-7}.

\bibitemdeclare{article}{method2018}
\bibitem{method2018}
\bibinfo{author}{Tarek \surnamestart Khaled\surnameend},
  \bibinfo{author}{Belaïd \surnamestart Benhamou\surnameend} \&
  \bibinfo{author}{Pierre \surnamestart Siegel\surnameend}
  (\bibinfo{year}{2018}): \emph{\bibinfo{title}{A new method for computing
  stable models in logic programming}}.
\newblock {\sl \bibinfo{journal}{Tools with Artificial Intelligence (ICTAI)}},
  pp. \bibinfo{pages}{800--807}, \doi{10.1109/ICTAI.2018.00125}.

\bibitemdeclare{article}{mushthofa2014}
\bibitem{mushthofa2014}
\bibinfo{author}{Mushthofa \surnamestart Mushthofa\surnameend},
  \bibinfo{author}{Gustavo \surnamestart Torres\surnameend},
  \bibinfo{author}{Yves \surnamestart Van~de Peer\surnameend},
  \bibinfo{author}{Kathleen \surnamestart Marchal\surnameend} \&
  \bibinfo{author}{Martine \surnamestart De~Cock\surnameend}
  (\bibinfo{year}{2014}): \emph{\bibinfo{title}{ASP-G: an ASP-based method for
  finding attractors in genetic regulatory networks}}.
\newblock {\sl \bibinfo{journal}{Bioinformatics}}
  \bibinfo{volume}{30}(\bibinfo{number}{21}), pp. \bibinfo{pages}{3086--3092},
  \doi{10.1093/bioinformatics/btu481}.

\end{thebibliography}
\bibliographystyle{eptcs}

\end{document}